\documentclass[letterpaper, 10 pt, conference]{ieeeconf}  
\usepackage{multirow}
\usepackage{makecell}
\IEEEoverridecommandlockouts                              

\usepackage{cite}
\usepackage{amsmath,amssymb,amsfonts}
\usepackage{algorithmic}
\usepackage{graphicx}
\usepackage{textcomp}
\def\BibTeX{{\rm B\kern-.05em{\sc i\kern-.025em b}\kern-.08em
    T\kern-.1667em\lower.7ex\hbox{E}\kern-.125emX}}
\usepackage{enumerate}
\usepackage{subcaption}
\usepackage[table,xcdraw]{xcolor}

\definecolor{PineGreen}{RGB}{0,153.0,0}
\definecolor{LightCyan}{rgb}{0.9,0.9,1}

\usepackage{array}

\usepackage{pifont}

\title{\LARGE \bf TOFFE - \underline{T}emporally-binned \underline{O}bject \underline{F}low \underline{f}rom \underline{E}vents for High-speed and Energy-Efficient Object Detection and Tracking}
\author{Adarsh Kumar Kosta, Amogh Joshi, Arjun Roy, Rohan Kumar Manna, Manish Nagaraj, and Kaushik Roy\\
Elmore Family School of Electrical and Computer Engineering, Purdue University\\
West Lafayette, IN 47907, USA\\{\tt \small $\{akosta, joshi157, roy208, rmanna, mnagara, kaushik\}@purdue.edu$}
}

\begin{document}

\maketitle
\thispagestyle{empty}
\pagestyle{empty}

\begin{abstract} 
Object detection and tracking is an essential perception task for enabling fully autonomous navigation in robotic systems. Edge robot systems such as small drones need to execute complex maneuvers at high-speeds with limited resources, which places strict constraints on the underlying algorithms and hardware. Traditionally, frame-based cameras are used for vision-based perception due to their rich spatial information and simplified synchronous sensing capabilities. However, obtaining detailed information across frames incurs high energy consumption and may not even be required. In addition, their low temporal resolution renders them ineffective in high-speed motion scenarios. Event-based cameras offer a biologically-inspired solution to this by capturing only changes in intensity levels at exceptionally high temporal resolution and low power consumption, making them ideal for high-speed motion scenarios. However, their asynchronous and sparse outputs are not natively suitable with conventional deep learning methods. In this work, we propose TOFFE, a lightweight hybrid framework for performing event-based object motion estimation (including pose, direction, and speed estimation), referred to as \textit{Object Flow}. TOFFE integrates bio-inspired Spiking Neural Networks (SNNs) and conventional Analog Neural Networks (ANNs), to efficiently process events at high temporal resolutions while being simple to train.  Additionally, we present a novel event-based synthetic dataset involving high-speed object motion to train TOFFE.
Our experimental results show that TOFFE achieves  5.7$\times$/8.3$\times$ reduction in energy consumption and 4.6$\times$/5.8$\times$ reduction in latency on an edge GPU(Jetson TX2)/hybrid hardware(Loihi-2 and Jetson TX2), compared to previous event-based object detection baselines.
\end{abstract}

\section{INTRODUCTION}
\label{Sec:Intro}

Artificial intelligence enabled robot systems of today draw significant inspiration from biological systems in an attempt to mimic them. Towards this, a critical task in achieving safe autonomous robot navigation is Object Detection and Tracking \cite{desouza2002survey} (further referred to as ODT), which involves identifying the location and motion behavior of objects in the environment and tracking them over time. Traditionally, object detection has been carried out using methods that require handcrafted features and rely on complex mathematical models \cite{viola2001rapid, dalal2005histograms, felzenszwalb2009object}. They involve scanning the input using sliding windows, extracting semantic features, using these features to find and classify objects \cite{zhao2019object}, as well as track them over time. As a result, these methods end up being slow and energy-intensive, making them unsuitable for high-speed motion scenarios.


In recent years, deep learning-based approaches have garnered substantial attention and achieved significant success. Convolutional Neural Networks (CNNs) are particularly effective at extracting relevant features from images and learning complex patterns that represent different object categories. CNN-based object detection architectures are broadly of two types: two-stage detectors and one-stage detectors. Two-stage detectors, such as R-CNN~\cite{rcnn}, Faster R-CNN~\cite{fasterrcnn}, SPPNet~\cite{sppnet} etc., incorporate a region proposal network, which generally leads to higher accuracy but at the cost of increased processing time. In contrast, one-stage detectors, like YOLO~\cite{redmon2016you}, RetinaNet~\cite{lin2017focal}, and SSD~\cite{liu2016ssd}, perform detection in a single pass, offering superior speed and real-time capabilities. More recently, transformer-based approaches, such as DETR~\cite{carion2020end} have surfaced, offering high performance improvements. Although, these methods significantly address the accuracy and latency limitations of traditional approaches, they remain highly power-intensive and still incur significant latency, making them unsuitable for high-speed, resource-constrained edge applications \cite{hu2023planning}. This underscores the necessity of exploring alternative, efficient pipelines — from sensors to algorithms to hardware — to meet the real-time demands of edge computing.


\begin{figure*}[t]
\centering
\includegraphics[width=\textwidth]{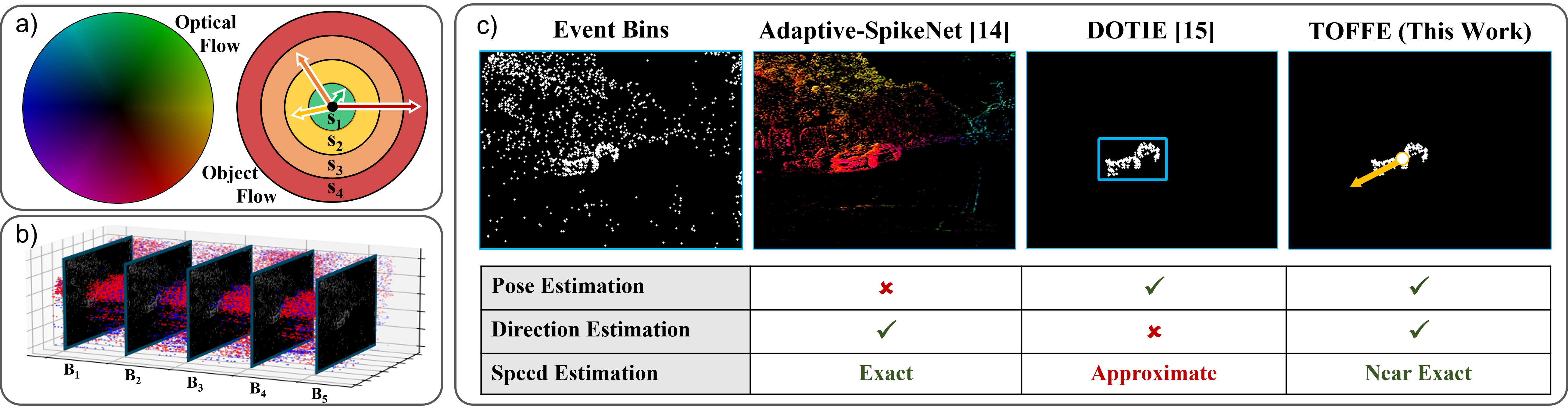}
\caption{a) Color wheels for optical flow and object flow. Optical flow has a continuous speed representation while object flow uses a discretized speed representation with arrows depicting the motion direction.  b)The raw stream of events in a given time window is discretized into five event-bins. These bins represent inputs at different timesteps when passed sequentially to an SNN. c) Comparison of TOFFE with Adaptive-spikenet \cite{adaptivespikenet} and DOTIE \cite{dotie} for object flow estimation. }
\label{fig:fig1}
\vspace{-5mm}
\end{figure*}

On the sensing end, frame-based cameras have been traditionally used for vision tasks. They synchronously capture rich photometric information, at each pixel with a lower temporal resolution, typically in the order of milliseconds. This makes them vulnerable to issues such as motion blur, and capturing dense spatial information across all pixels leads them to have a high energy consumption and low dynamic range due to bandwidth limitations. In contrast, living organisms such as winged insects equipped with biological eyes, can perform complex, high-speed maneuvers in dense environments using only visual cues from their surroundings \cite{flymotion2010, honeybee1996, serres2017optic}. Unlike frame-based cameras, the biological eye operates in an asynchronous manner, gathering and processing only the changes in visual information and leading to a fast and highly efficient sensing mechanism. Coupled with the efficient computations in the biological brain, enables the agile behaviour demonstrated by these organisms. Inspired by the biological retina~\cite{mahowald1994silicon}, event-based cameras~\cite{dvs1, dvs2} have been proposed. These cameras sample the log intensity changes at each discrete pixel asynchronously and independently, albeit with reduced photometric detail. Any change in the log intensity ($I$) over a specified threshold ($\theta$) is recorded as a discrete binary event at that pixel location, (i.e., $\|\log(I_{t+1}) - \log(I_{t})\| \geq \theta$). This approach not only minimizes energy consumption and reduces motion blur but also enables operation across a wide dynamic range. These features position event-based cameras as ideal candidates for high-speed, low-energy autonomous navigation systems.

Nevertheless, the sparse and asynchronous stream of events is not compatible with conventional Artificial Neural Networks (ANNs) as they are predominantly designed for synchronous frame-based inputs. In contrast, bio-inspired models such as Spiking Neural Networks (SNNs) can effectively handle event data due to their inherent sparse, asynchronous and event-driven computations similar to the brain. However, SNNs are difficult to train due to several challenges such as non-differentiable activation function, ``vanishing spikes'' \cite{rathi2020enabling} etc. Thus, a hybrid SNN-ANN architecture exploiting the complementary benefits of each can potentially lead to smaller, energy efficient models suitable for real-time processing in edge autonomous systems. 

To accomplish the ODT task, an autonomous robot should be able to estimate three features of the objects present in the environment, which we collectively term as \textit{Object Flow}: 
(a) \textit{pose} of the object, (b) \textit{direction} of motion of the object and, (c) \textit{speed} the object. In essence, object flow can be understood as optical flow for objects with discretized speed and added pose information, as depicted in Fig. \ref{fig:fig1}a. 
Object flow can be broadly subdivided into two components - (a) object flow pose and direction (OFPD) estimation for detection, and (b) object flow speed estimation (OFS) to aid tracking. Intuitively, both these sub-tasks operate on separate information modalities - detection relies more on the spatial information contained in the pixels while tracking requires the temporal information provided by inter-frame motion. To that effect, leveraging an architectural combination of SNNs (for OFS) and ANNs (for OFPD) can potentially lead to a high-performance yet lightweight solution capable of meeting the energy and latency requirements of edge systems even in high-speed motion scenarios. 

The main contributions of this paper are as follows:
\begin{enumerate}
    \item We present TOFFE, a novel lightweight hybrid SNN-ANN algorithm that estimates the pose, direction, and speed for objects in the field of view, termed as \textit{Object Flow}. It provides a spatially sparse and speed discretized alternative to traditional optical flow.
    \item We generate and provide a novel high-speed synthetic dataset used to train TOFFE in a supervised fashion. This dataset consists of sequences with objects moving at different speeds in a variety of trajectories.
    \item We showcase the performance improvements of TOFFE compared to state-of-the art lightweight baselines.
    \item We demonstrate the energy and latency benefits of TOFFE on traditional GPUs as well as on a hybrid of traditional and neuromorphic hardware.
\end{enumerate}

\section{Related Work}
\label{sec:Lit_survey}


\paragraph{Optical Flow Estimation} An important aspect of autonomous navigation systems is estimating the motion of the objects in its surroundings. Optical flow offers a generic solution to this by estimating motion for every input pixel. Many ANN-based solutions, \cite{opticalflowcnn1, opticalflowcnn2, evflownet} have been widely used for optical flow estimation using frame-based camera outputs. For event-based inputs, works such as \cite{opticalflowsnn1, opticalflowsnn2, opticalflowsnn3} explored various ways of leveraging SNNs to effectvely utilize the spatiotemporal information provided by event-based cameras compute optical flow. However, these works fail to scale to larger and more realistic datasets. To overcome this issue, authors in \cite{spikeflownet} utilized a hybrid architecture consisting of an SNN encoder and an ANN decoder to combat the poor performance of deep SNNs. Following this, Adaptive-SpikeNet \cite{adaptivespikenet} employed a fully SNN architecture with learnable spiking neuronal dynamics outperforming all the previous works, both in terms of performance and efficiency. However, in regard to the ODT task, optical flow provides direction and speed estimates for individual pixels, with no information about object pose. In addition, these high-precision pixel-wise motion estimates may not be suitable for effective object tracking as a single direction and speed estimate per object is required. Relaxing the constraints to cater to estimating ``\textit{Object Flow}'' can lead to even more efficient implementations. 

 \begin{figure}[t]
\centering
\includegraphics[width=\columnwidth]{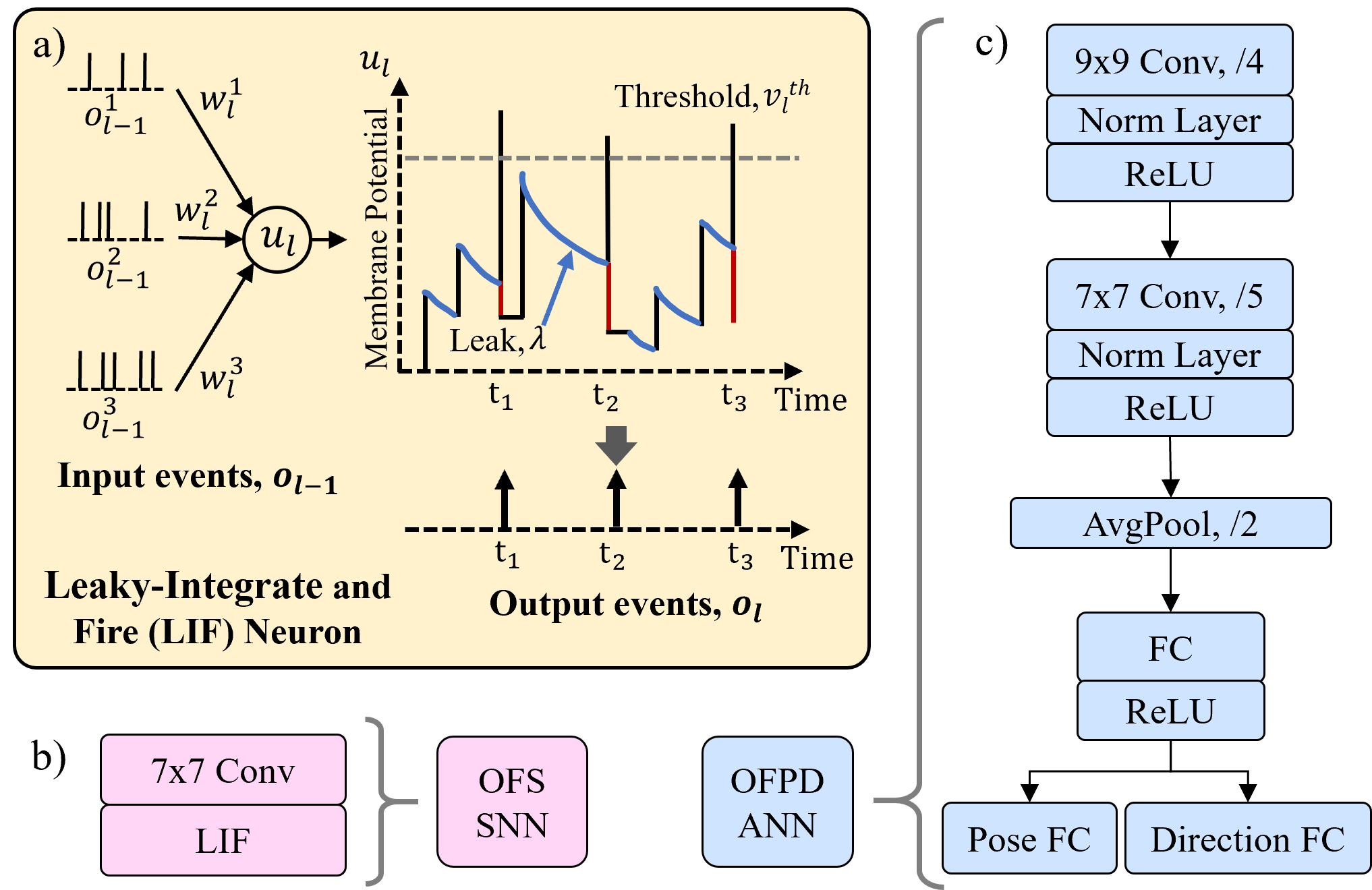}
\caption{a) A Leaky-Integrate and Fire (LIF) spiking neuron with firing threshold ($v^{th}_l$) and leak factor ($\lambda$). b) SNN-based OFS (Object Flow Speed) block. c) ANN-based OFPD (Object Flow Pose and Direction) block.}
\label{fig:lif_ofs_ofpd}
\vspace{-5mm}
\end{figure}

\paragraph{Object Detection and Tracking}
For traditional frame-based camera outputs, the abundance of rich photometric features makes it simpler to detect and track objects. Several works originally based on handcrafted features and mathematical models  \cite{viola2001rapid, dalal2005histograms, felzenszwalb2009object} and more recent works based on deep learning \cite{rcnn, fasterrcnn, sppnet, redmon2016you, lin2017focal, liu2016ssd} have been proposed. Events on the other hand, while temporally rich, lack such photometric features. In recent years, there has been an emerging interest to focus on object detection and tracking using events. Early works used blob detectors to identify patterns inherently present in events \cite{earlyfeat1, earlyfeat2, blobdet}. These algorithms, however, only operated in simple scenarios. Some algorithms, such as \cite{EED:mitrokhin2018event} also utilized motion compensation techniques to compensate for the system's ego-motion to allow for object detection in more complex scenarios. To further improve accuracy, events were aggregated during a time duration to form ``frames" in \cite{cnn1, cnn2, cnn3, cnn4, cnn5:YOLE}. These ``frames" were then fed into traditional ANNs to detect the object features. A more recent work, DOTIE \cite{dotie} tries to optimally use events and reduce the compute and latency overheads caused by ANNs. It employs a extremely lightweight spiking architecture that can separate events belonging to an object based on the speed of motion of the object. However, following this, a spatial clustering algorithm (DBSCAN \cite{dbscan}) is employed to detect the object pose, which significantly slows down inference and incurs high computational cost. It also lacks direction estimation, making tracking challenging.

Evaluating closely related works, Adaptive-SpikeNet \cite{adaptivespikenet} can perform direction and speed estimation at an elevated compute cost, while DOTIE \cite{dotie} can only perform pose and approximate speed estimation, as shown in Fig. \ref{fig:fig1}c.
While both works have shown the feasibility of SNNs with event-based cameras, they fall short of meeting the latency requirements for high-speed motion scenarios due to a relatively large architecture in \cite{adaptivespikenet} and the use of conventional clustering in \cite{dotie}. With these in mind, TOFFE proposes an energy-efficient framework that can carry out the entire object flow estimation pipeline with latency and energy requirements within the resource budget at the edge.

\section{Methodology} \label{Sec:Methodology}

\subsection{Input Representation}
Event-based cameras generate data in the Address Event Representation (AER) format comprising a tuple $\{x, y, t, p\}$, with $(x,y)$ representing pixel locations, ($t$) representing the camera timestamp and ($p$) representing the (ON/OFF) polarity of the intensity change. In this work we utilize a discretized event volume representation similar to \cite{adaptivespikenet}, where raw events in a given time window ($dt$) are mapped into finite number of event bins ($B=5$) as depicted by $B_1$, $B_2$, $B_3$, $B_4$ and $B_5$ in Fig. \ref{fig:fig1}b. These event bins are treated as inputs over timesteps when passed to an SNN. 

\subsection{Spiking Neuron Model}
We utilize the leaky integrate-and-fire (LIF) neuron model \cite{abbott1999lapicque}, chosen for its robustness for information storage and retrieval, alongside its simplicity compared to conventional recurrent neural networks (RNNs), allowing for a lightweight architectural framework without compromising performance. The LIF neuron integrates information over time through the accumulation of membrane potential ($u$), while concurrently enabling regulated forgetting through a leakage parameter ($\lambda$). The dynamic behavior of the LIF neuron is depicted in Fig. \ref{fig:lif_ofs_ofpd}a and is characterized as follows:

\begin{equation}
    \boldsymbol{u}_l^t = \lambda_l\boldsymbol{u}_l^{t-1}  +  \boldsymbol{W}_lo_{l-1}^t - v^{th}_l \boldsymbol{o}_l^{t-1} \\ \tag{1}
    \label{eq1}
\end{equation}
where for the layer $l$, $ \boldsymbol{u}_l^t$ represents the membrane potential at timestep $t$, $ \boldsymbol{W}_l$ represents the synaptic weights connecting to the previous layer $l-1$, $o_l$ represents the binary output spike, $v^{th}_l$ depicts the firing threshold and $\lambda_l$ represents the leak factor.
Output spikes are generated according to the following equation at each timestep:
\begin{equation}
     \boldsymbol{z}_l^{t} =  \boldsymbol{u}_l^{t}/v^{th}_l - 1, \; \; \; \;  
     \boldsymbol{o}_l^{t} =\begin{cases}
               1, & \text{if~$ \boldsymbol{z}_l^{t}>0$}\\
               0, & \text{otherwise}
            \end{cases} \tag{2}
    \label{eq2}
\end{equation}
The threshold ($v^{th}$) dictates the average duration of input integration, while the leak ($\lambda$) regulates the retention of membrane potential between timesteps. Inspired by \cite{rathi2021diet, adaptivespikenet}, the $v^{th}$ and $\lambda$ parameters are kept trainable to enhance the learning capacity of the LIF neuron resulting in lightweight architecture that can handle complex tasks. 

\begin{figure*}[t]
\centering
\includegraphics[width=0.85\textwidth]{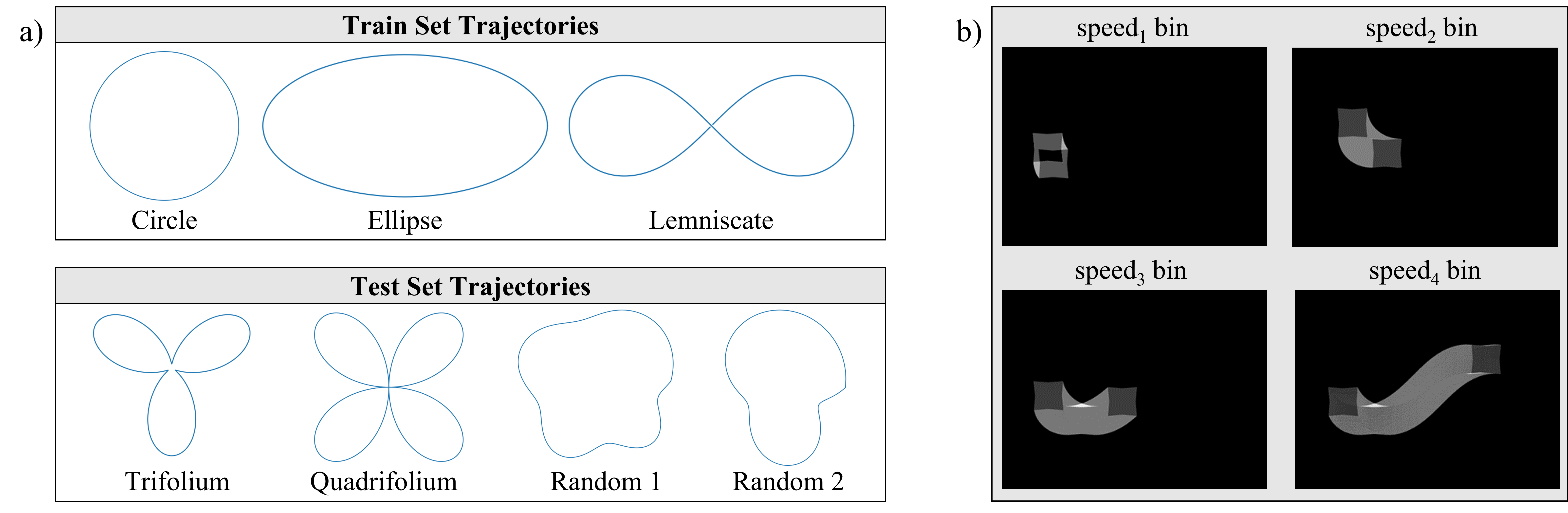}
\caption{a) Train and Test set trajectories for for TOFFE dataset. b)  Accumulated event-bin images corresponding to speeds - 1 to 4 for lemnniscate trajectory rendered at 30FPS.}
\label{fig:trajectories_and_renders}
\vspace{-5mm}
\end{figure*}

\begin{figure}[ht]
\centering
\includegraphics[width=0.8\columnwidth]{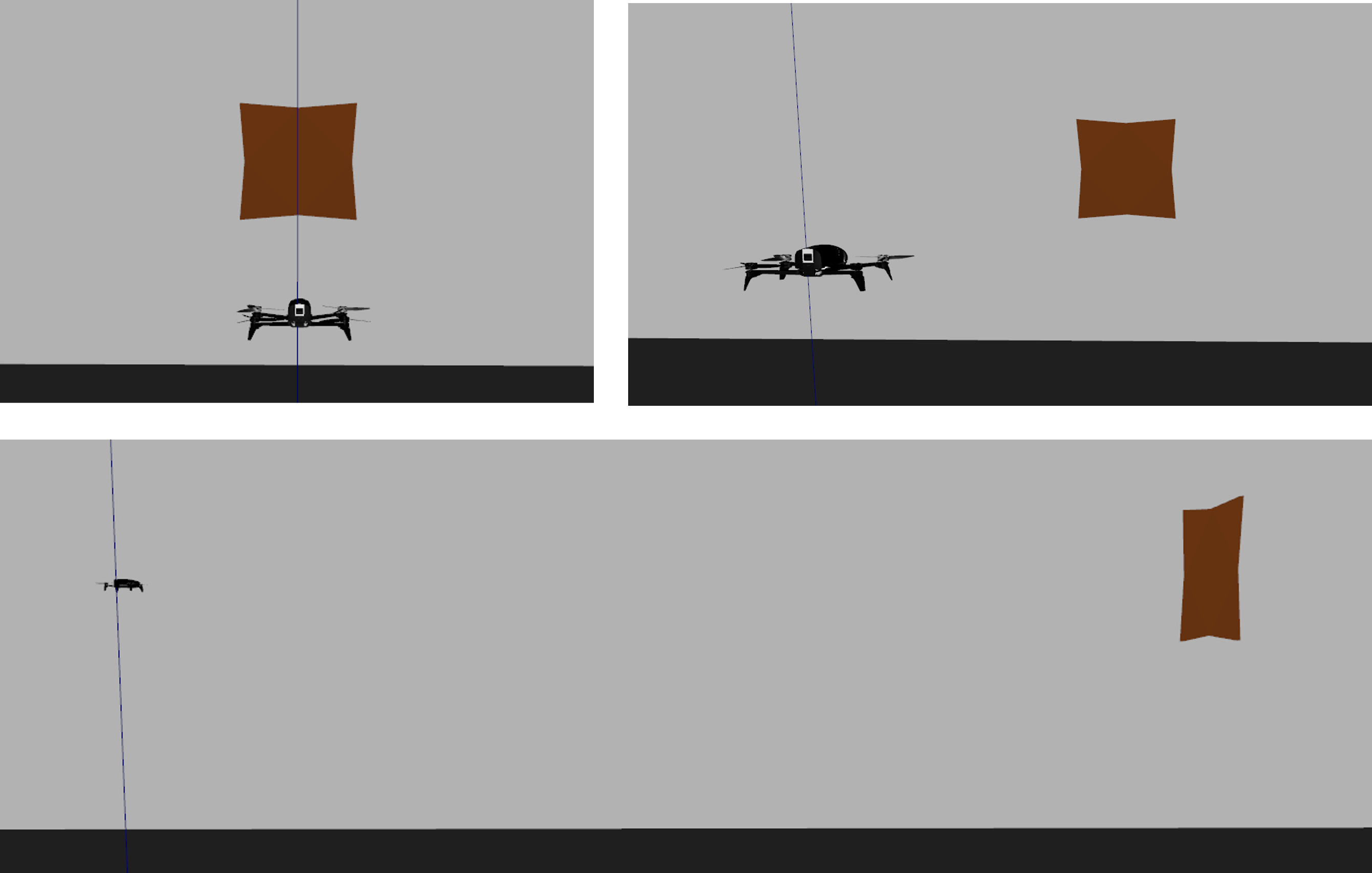}
\caption{Data collection setup in Gazebo.}
\label{fig:simulator}
\vspace{-5mm}
\end{figure}

\subsection{TOFFE Dataset}


Well-known existing datasets for robot navigation include frame-based Waymo open \cite{waymo} and nuScenes datasets \cite{nuscenes} for object detection and tracking, and event-based MVSEC \cite{mvsec} and DSEC \cite{dsec} datasets for optical flow estimation. All these datasets involve data collection in the real-world for mostly outdoor driving and some indoor drone flying scenarios. Although quite comprehensive and accurate in terms of the groundtruths offered, they lack data as well as groundtruth information for high-speed motion scenarios. On the sensing end, this can be fundamentally attributed to the usage of low temporal resolution sensors such as frame-based cameras, LIDARs, RADARs etc \cite{waymo, nuscenes}. In addition, it is quite challenging to reliably collect high speed motion data in an uncontrolled real-world enviornment. Some sensors such as event-based cameras aid in capturing high-speed motion in some datasets \cite{mvsec, dsec}, however the groundtruth generation for this data is still capped at a low temporal resolution due to the limitations associated with other sensors.



This incites the need to switch to synthetic data generation to address the above challenges. Synthetic datasets \cite{fedora} can be collected in a controlled simulation environment with adjustable simulation rate to capture high-speed information. Thus, sensing and generating accurate groundtruth information at a very high rate is possible. Therefore, as part of this work, we develop a synthetic event-vision dataset, called TOFFE dataset to perform the object flow task. The dataset records frames and depth information at $30$ $samples/s$ in addition to events and 6-dof pose inforamtion at $20000$ $samples/s$. It also provides accurate ground truth object pose and velocity at a rate of $20000$ $samples/s$, much faster than real-time and ideal for high-speed motion scenarios.

\begin{table}[ht]
    \centering
    \renewcommand*{\arraystretch}{1.1}
    \caption{Specifications of the Simulated Sensors} 
    \label{tab:sensor_spec}
    \begin{tabular}{|c|c|}
    \hline
        Sensor & Specifications \\
        \hline
        \multirow{2}{*}{\makecell{VI-Sensor \\ (Frame Camera)}} & 1440x1080 pixels @30 fps \\
         & FoV: 60$^\circ$ horizontal\\
        \hline
        \multirow{3}{*}{\makecell{RealSense Depth\\Sensor}} & 1440x1080 pixels @30 fps \\ & Sensing limits: [0.1, 30]m \\
         & FoV: 60$^\circ$ horizontal\\
        \hline
        \multirow{2}{*}{\makecell{DAVIS Sensor \\ (Event Camera)}} & 640x480 pixels @ 20K samples/s\\
         & FoV: 60$^\circ$ horizontal\\
        \hline
        \multirow{2}{*}{\makecell{Pose Logger\\(mounted on object)}} & 6 DoF Pose Data  \\ & @ 20000 samples/s \\
        \hline
    \end{tabular}
\end{table}

\begin{table}[ht]
    \centering
    \renewcommand*{\arraystretch}{1.1}
    \caption{Speed ranges for the four speed bins}
    \label{tab:sp_range}
    \begin{tabular}{|c|c|c|}
    \hline
        \multirow{2}{*}{Speed Bin} & \multicolumn{2}{c|}{Speed Range} \\
        \cline{2-3}
         & Min Speed (m/s) & Max Speed (m/s) \\
        \hline
        1 & 1 & 18 \\
        2 & 18 & 42 \\
        3 & 42 & 84 \\
        4 & 84 & 500 \\
        \hline
    \end{tabular}
\vspace{-2mm}
\end{table}

TOFFE dataset is recorded in the Gazebo simulator with a  sensor suite mounted on a stationary drone, as shown in Fig. \ref{fig:simulator} to observe and record moving objects. Table \ref{tab:sensor_spec} lists the specifications of our sensor suite that incorporates a VI-sensor, a depth camera, a DAVIS camera \cite{dvs240} (events + frames), and a pose-logger. A true DVS sensor is asynchronous in nature having a sub-microsecond ($\mu s$) sampling time. However, due to machine and simulation restrictions, the simulated DVS sensor used in this work (based on \cite{dvs240}) is limited to a minimum yet significantly low inter-event interval of $50\mu s$ ($20000$ $samples/s$). Four object variations are considered: square, circle, diamond and star moving at various speeds. Fig. \ref{fig:trajectories_and_renders}a shows the train and test motion trajectories that are parameterized in cylindrical coordinates with same lap time. Such parameterization results in varying speeds at different locations within a trajectory sequence, which allows to capture a speed range and is beneficial for the ODT network's speed-separation capabilities. These speeds are binned into four exclusive speed bins, whose ranges are as shown in Table \ref{tab:sp_range}. Fig. \ref{fig:trajectories_and_renders}b showcases an frame constructed from accumulated events for a square object moving in a lemniscate trajectory at speeds corresponding to the four speed bins, rendered at $30$ $samples/s$.  After a sequence is recorded, we post-process it to generate object pose and velocity groundtruths using information from the pose logger.

\begin{figure*}[t]
\centering
\includegraphics[width=\textwidth]{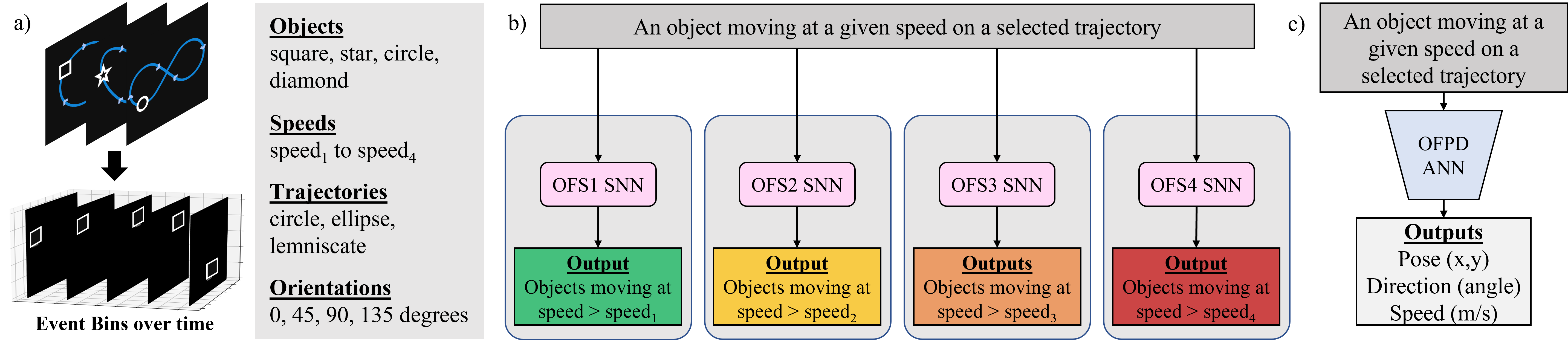}
\caption{TOFFE Training with four speeds (N=4). TOFFE is trained on many different objects, trajectories and speeds. TOFFE learns the pose, direction and speed of objects in the image plane.}
\label{fig:architecture_training}
\vspace{-5mm}
\end{figure*}

While the maximum programmed object speed was $144m/s$, we set the upper speed threshold of bin-4 to $500m/s$ to account for stray events that may be perceived by the network as having a higher speed. For the training set, we collected sequences by varying object speed in the four speed bins, the motion direction (clockwise/anti-clockwise), and trajectory orientation ($0^\circ$, $45^\circ$, $90^\circ$ and $145^\circ$) across all three train set trajectories, and all four object shapes. The test set was collected in a similar manner, with different trajectories but with constant trajectory orientation of $0^\circ$.



\subsection{TOFFE Architecture}
A naïve architecture for the object flow task would be a monolithic Analog Neural Network (ANN) or Spiking Neural Network (SNN). Training such a network to learn the object flow task would require the network to classify objects based on speed and differentiate between objects corresponding to different speed bins to predict their pose and direction. Ensuring convergence in the training of such a network is a challenging problem. Additionally, such a network is bound to be large in size owing to the number of different tasks it must learn.

As discussed earlier, computing object flow (OF) for the ODT task can be divided into two-distinct subproblems: discriminating between events based on speed and predicting the pose and direction of the discriminated events. The former can be solved mostly using temporal information while the latter would require both spatial and temporal information. Intuitively, networks better suited to learning temporal information are ideal candidates for solving the former problem, while the latter would require networks that can handle spatial information. Therefore, we choose a best-of-both-worlds approach, using a learnable DOTIE-like \cite{dotie} SNN for speed separation (approximate speed estimation) (OFS) and designing a lightweight ANN for pose and direction estimation (OFPD). Note that, DOTIE \cite{dotie} did not use any learning and had its network parameters manually tuned and set apriori. The OFS and OFPD pipelines are described below:

\subsubsection{Object Flow Speed (OFS)}
This is the speed separation portion of the TOFFE architecture, which incorporates a modified version of the DOTIE SNN. DOTIE \cite{dotie} uses a single-layer fine-tuned SNN to achieve speed separation for a single speed. DOTIE outputs events only for objects moving at a speed greater than the set speed. However, this approach is not very scalable for real-life scenarios where multiple speeds need to be separated using multiple speed bins. Therefore, in this work, we take a single-layer SNN and train it along with its neuronal dynamics (threshold and leak parameters) to enable separating out multiple speeds. We consider separation into multiple different speed ranges (or bins) in this work using independently trained OFS models. In addition to this, the usage of LIF neuron allows for noise filtering in noisy environments, making the system robust to external noise (as shown later in noise ablations). The architecture of OFS SNN is shown in Fig. \ref{fig:lif_ofs_ofpd}b.

\subsubsection{Object Flow Pose and Direction (OFPD)}
The speed separated events for speeds within a speed bin are passed on to the OFPD ANN that performs pose and direction estimation on them. The OFPD ANN consists of a two layered convolutional network followed by a fully-connected (FC) layer. The ouput of the FC layer splits into the pose and direction heads, that estimated the pose and direction values of the observed object. The pose head outputs the location of the center of the object as a $(x,y)$ tuple, while the direction head outputs the angle in radians corresponding to the motion direction of the object. The architecture of OFPD ANN is shown in Fig. \ref{fig:lif_ofs_ofpd}c.

\subsection{TOFFE Training}
The OFS and OFPD architectures operate in tandem with each other to perform object flow estimation over multiple speeds as shown in Fig. \ref{fig:architecture_training}. For OFS training on each of the speed bins, a corresponding OFS architecture is trained and used to identify objects with corresponding and higher speeds of motion. Thus, $OFS_1$ outputs events for objects moving at speed-1 or greater, $OFS_2$ outputs evetns for objects moving at speed-2 or greater and so on. Thus, a post-processing step is required during inference to exclusively separate OFS outputs and will be discussed in the next section.
The OFPD model is trained on events inputs consisting of moving objects from all speed bins, albeit a single speed per input and thus learns to estimate pose and direction of objects irrespective of their speed of motion. A single OFPD model is thus trained in contrast to multiple OFS models (one for each speed bin). Both, the OFS and OFPD training is carried out in a supervised manner and therefore rely heavily on the accuracy of ground truth available for the corresponding dataset used.

\begin{figure}[t]
\centering
\includegraphics[width=\columnwidth]{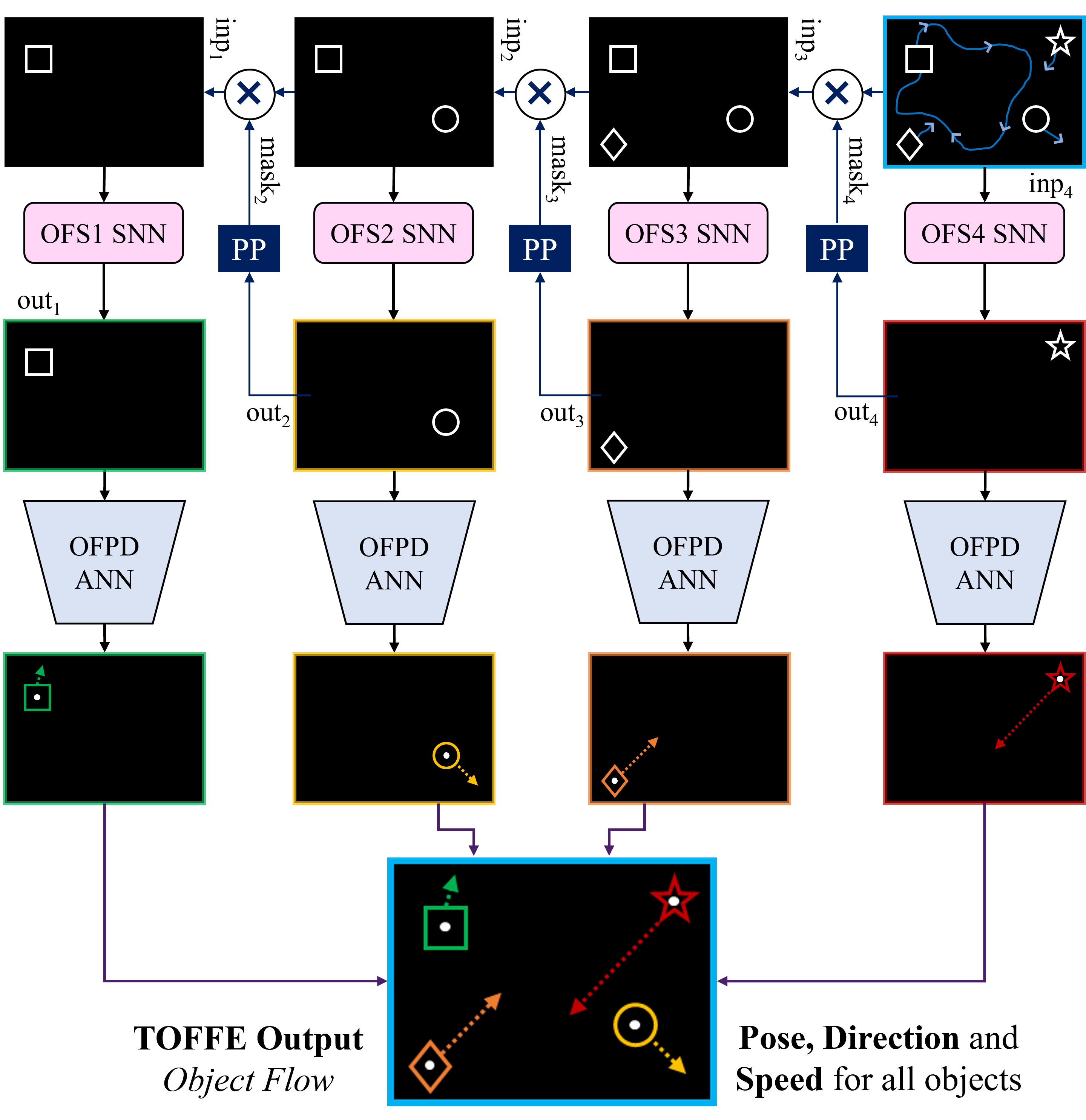}
\caption{TOFFE Inference with four speeds (N=4). TOFFE can be fed a sequence with many objects moving independently and estimate the pose, direction, and speed of each individually. This method is shape-agnostic and detects objects based on their speed.}
\label{fig:architecture_inference}
\vspace{-5mm}
\end{figure}

\subsection{TOFFE Inference}
During inference, a raw event stream of objects moving at different speeds is first converted into discretized event bins ($B$ bins in $T$ duration). The OFS level processes these inputs in a sequential manner, going from highest speed bin ($N$) to lowest ($1$). This is because a spiking neuron acts like a high pass filter and thus $OFS_K$ will generate outputs for objects moving at not just $speed_k$ but at $speed_k$ or higher. Events belonging to $speed_k$ and higher, constitute the $OFS_k$ output and are consequently removed from inputs to the remaining $OFS$ models. This is done through a post-processing (PP) step by computing an output mask using standard morphological operations (closing and inversion) and masking out the input using this mask, as shown below.

\begin{equation}
    mask_{k} = \textit{Invert( Close(}out_k\textit{) )}
    \tag{3}
    \label{eq3}
\end{equation}
\begin{equation}
    inp_{k-1} = inp_{k}*mask_{k}
    \tag{4}
    \label{eq4}
\end{equation}

The \textit{Closing} operation on the output of $OFS_k$  ($out_k$) incorporates \textit{dilation} followed by \textit{erosion} operations. This allows to increase the density of events corresponding to $speed_k$ and fill in any gaps to remove discontinuities. This is followed by a binary inversion operation to generate $mask_k$. The binary mask is multiplied with the input to the $OFS_K$ model ($inp_k$) to generate the input to generate the input to the  $OFS_{K-1}$ model ($inp_{k-1}$). Thus, the $OFS_{K-1}$ model receives the processed events with events corresponding to $speed$ $>$ $speed_k$ and above removed. A kernel size of $5$ is used for all morphological operations.

The obtained OFS outputs ($out_k$) with $k\in{1,2,..,N}$ are passed through N-copies of the originally trained OFPD ANN in parallel. Corresponding OFPD outputs ($1-N$) estimate the pose and direction for object in each speed bin and are stitched together to obtain a composite representation showing all objects in the original input with their approximate speeds, pose, and motion directions Fig. \ref{fig:architecture_inference} shows the inference pipeline for TOFFE.

\section{Results}
\label{Sec:Results}




\subsection{Object Flow Accuracy}
The performance of TOFFE is evaluated in terms of the errors obtained for Object flow subtasks. These include errors for pose estimate (pixE), direction estimate (dirE) and, speed estimate (speedE), calculated as mean absolute errors with respect to the groundtruth. TOFFE obtains noteworthy error values, that allow effective tracking of the objects moving at speeds corresponding to all speed bins, as depicted in Table \ref{tab:results}. We see that the choice of $dt$ governs the detection and tracking accuracy that can be obtained. At smaller $dt$ values, the events comprising the input are sparse and lack structure, leading to resulting in higher errors direction and speed estimates. This can be seen for $dt200$ with high dirE of $24.4^{\circ}$ and speedE of $15.6 m/s$, making it unsuitable for tracking. On the other hand, higher $dt$ values incorporate more events leading to a denser representation. This accumulation of events over a longer time interval can make the object motion appear jittery by increasing the sampling period between successive inputs and thereby increasing dirE and speedE. The increased events can also alter the thickness of the object boundary disproportionately during motion, adding error to the pose estimate (pixE). This is evident from the results corresponding to $dt5000$ in Table \ref{tab:results}.
We observe that $dt500$ and $dt1000$ turn out to be the optimal event time windows for constructing the input event bins for TOFFE. 

\begin{table}[h]
    \centering
    \setlength{\tabcolsep}{6pt}
    \renewcommand*{\arraystretch}{1.2}
    \caption{TOFFE results for different $dt$ values}
    \label{tab:results}
    \begin{tabular}{|c|c|c|c|}
    \hline $dt$ & pixE (pixels) & dirE (degrees) & speedE (m/s)\\
    \hline
    $dt5000$ & 14.135 & 25.842 & 19.948\\
    $dt2000$ & 8.406 & 14.345 & 16.677\\
    $dt1000$ & 6.626 & \underline{13.817} & \underline{11.867}\\
    $dt500$ & \textbf{5.355} & \textbf{10.769} & \textbf{10.649}\\
    $dt200$ & \underline{5.404} & 24.411 & 15.632\\
    \hline
    \end{tabular}
\vspace{-2mm}
\end{table}
 
\subsection{Computational Efficiency}
To demonstrate the computational efficiency of TOFFE we investigate two key metrics: \textit{(1) dynamic energy per inference (E)} and \textit{(2) inference latency (L)} on traditional von-Neumann and neuromorphic hardware. To calculate the dynamic energy per inference we subtract the idle power consumption of the hardware, when no programs are running, from the power consumed when only running the models. The inference latency is simply the time taken to perform one inference. On neuromorphic hardware, this depends on the number of bins per inference ($B=5$) and the time taken per bin $T_{ts}$. Latency is then computed as $L=T_{ts}*B$.

\begin{table*}[!htp]
    \centering
    \setlength{\tabcolsep}{6.5pt}
    \renewcommand*{\arraystretch}{1.2}
    \begin{tabular}{|c|c|c|c|c|c|c|c|c|c|}
    \hline
    \multirow{3}{*}{Model} & \multirow{3}{*}{Output} & \multicolumn{2}{|c|}{CPU} & \multicolumn{2}{|c|}{GPU} & \multicolumn{2}{|c|}{Edge GPU} & \multicolumn{2}{|c|}{Hybrid Hardware} \\
    & & \multicolumn{2}{|c|}{i9-12700KF} & \multicolumn{2}{|c|}{RTX 3090-Ti} & \multicolumn{2}{|c|}{Jetson TX2} & \multicolumn{2}{|c|}{Loihi2/Jetson-TX2} \\ \cline{3-10} 
    & & E($mJ$)$\downarrow$ & L($ms$)$\downarrow$ & E($mJ$)$\downarrow$ & L($ms$)$\downarrow$ & E($mJ$)$\downarrow$ & L($ms$)$\downarrow$ & E($mJ$)$\downarrow$ & L($ms$)$\downarrow$ \\ \hline \hline
    Adaptive-SpikeNet (Pico) \cite{adaptivespikenet} & Optical Flow & 1029.1 & 24.04 & 417.2 & 4.14 & 89.2 & 41.84 & - & - \\ \hline
    DOTIE (w/o Clustering) \cite{dotie} & Speed Separation & 28.8 & 0.62 & 31.1 & 0.34 & 4.6 & 5.95 & 0.083 & 0.29 \\
    DOTIE (with Clustering) \cite{dotie} & Object Detection & 558.0 & 16.8 & 426.9 & 4.50 & 74.2 & 93.46 & 69.6 & 88.67 \\ \hline
    \rowcolor{LightCyan} \textbf{TOFFE (Ours)} & Object Flow & 92.5 & 2.17 & 123.0 & 1.34 & 13.0 & 20.12 & 8.38 & 15.33 \\ 
    \hline
    \end{tabular}
\captionof{table}{Dynamic Energy per Inference [E] ($mJ$) and Inference latency [L] ($ms$) results of SOTA lightweight SNN approaches for object detection, optical flow and combined outputs. We use DOTIE (with Clustering) as a baseline as it comprises of a single-layer SNN with LIF neurons for speed separation (similar to our OFS SNN), followed by a clustering method for object pose estimation (similar to our OFPD ANN).}
\label{tab:energy_and_latency}
\end{table*}


All measurements on Loihi-2 are conducted using Lava 0.5.1, Intel's software framework for developing neuromorphic algorithms, on an Oheo Gulch board containing a single Loihi-2 chip. This chip is capable of supporting up to 1,000,000 fully-programmable neurons. For comparative purposes, measurements on the CPU are made using an Intel i9-12700KF processor (5.0 GHz), and measurements on the GPU are conducted with both a high-end NVIDIA RTX 3090 (24GB GDDR6X) and a power-efficient 256-core NVIDIA Pascal GPU in an NVIDIA Jetson-TX2 edge device.

While neuromorphic hardware offers significant advantages, it also has some limitations. For example, model size constraints are imposed by the total number of neurons available on the Oheo Gulch chip. Additionally, the Lava framework currently lacks support for certain complex operations, preventing us from running models such as Adaptive-SpikeNet and DOTIE (with clustering) on Loihi-2. Adaptive-SpikeNet exceeds the memory capacity of a single Oheo Gulch chip, and Lava does not yet support operations like skip connections. Similarly, DOTIE (with clustering) involves an iterative clustering algorithm that is not yet supported on Loihi-2. As the Lava framework evolves and neuromorphic hardware advances, we anticipate being able to run such models. In addition, the OFPD ANN in TOFFE requires a traditional CPU or GPU.  Thus, we employ a hybrid hardware setup combining Loihi-2 and the Jetson-TX2 for running the DOTIE and TOFFE models.

The results for dynamic energy per inference and inference latency are presented in Table \ref{tab:energy_and_latency}. DOTIE (without clustering) consists of a single convolutional operator with LIF neurons (similar to the OFS SNN in TOFFE), to perform speed-based separation of events. Note that, DOTIE (without clustering) is an incomplete implementation and is only used to highlight the energy efficiency and inference speed achievable with an SNN on Loihi-2 compared to traditional hardware. 

When evaluated on traditional hardware, TOFFE demonstrates significantly lower energy consumption and inference latency than both Adaptive-SpikeNet \cite{adaptivespikenet} and DOTIE (with clustering) \cite{dotie}, across all hardware types (CPU, GPU, and edge GPU). When running on a hybrid hardware setup of Loihi-2 and Jetson-TX2, we observe that TOFFE consumes substantially less energy than DOTIE (with clustering) and exhibits much lower inference latency. Comparing TOFFE implementations across traditional and hybrid hardware reveals that energy consumption is significantly reduced on hybrid hardware, as the SNN component is offloaded to neuromorphic hardware. In terms of inference latency, the hybrid hardware performs better than the edge GPU but worse than the high-end CPU and GPU. In edge systems, where the Jetson TX2 is more feasible than the RTX 3090, hybrid hardware outperforms traditional hardware in both energy consumption and inference rate. Infact, TOFFE's inference rate on hybrid hardware of approximately 65 frames per second is likely sufficient for most real-world applications. This inference rate is comparable to running DOTIE (with clustering) and Adaptive-SpikeNet on a high-end CPU, which consumes approximately three orders of magnitude more energy than TOFFE on the hybrid Loihi-2 and Jetson-TX2 hardware.

\section{Conclusion}
In this work, we explore and evaluate the challenges associated with object detection and tracking in high-speed motion scenarios. We highlight the limitations of current systems and emphasize the need for a lightweight sensor-algorithm-hardware pipeline. We utilize high temporal resolution event data and propose TOFFE, a trainable, lightweight hybrid SNN-ANN architecture designed for accurate object flow estimation, encompassing pose and motion estimation. Our results demonstrate that TOFFE achieves performance comparable to state-of-the-art object detection and optical flow methods while operating with significantly lower energy consumption and latency. Furthermore, its use of bio-inspired spiking neurons provides robustness to noisy inputs and makes it particularly suitable for low-power, low-latency edge applications involving high-speed motion.

\bibliography{main.bib}
\bibliographystyle{ieeetr}

\end{document}